%% file: main.tex
\documentclass[letterpaper, 10 pt, conference]{ieeeconf}  

\IEEEoverridecommandlockouts                              

\usepackage{graphics} 
\usepackage{mathptmx} 
\usepackage{amsmath} 
\usepackage{amssymb}  

\usepackage{balance}
\usepackage{comment}

\usepackage[hidelinks]{hyperref}
\usepackage{cleveref}
\usepackage{bm}
\usepackage[ruled,vlined]{algorithm2e}

\usepackage{booktabs}
\usepackage{adjustbox}
\usepackage[table]{xcolor}

\usepackage{caption}
\title{\LARGE \bf
A Nonlinear Complementarity Formulation of Frictional Contact \\ for Implicit Material Point Method
}

\author{Etienne Ménager$^*$ and Justin Carpentier$^*$
\thanks{$^{*}$Both authors are in Inria and Département d’Informatique de l’École Normale Supérieure, PSL Research University in Paris, 75013 Paris, France.
{\tt\small email: etienne.menager@inria.fr, justin.carpentier@inria.fr}}}%

\crefname{section}{Sec.}{Secs.}
\Crefname{section}{Section}{Sections}
\crefname{figure}{Fig.}{Figs.}
\crefname{equation}{Eq.}{Eqs.}
\crefname{table}{Table}{Tables}
\Crefname{figure}{Figure}{Figures}
\crefname{algocf}{Alg.}{Algs.}
\Crefname{algocf}{Algorithm}{Algorithms}

\input{macros}

\begin{document}

\maketitle
\thispagestyle{empty}
\pagestyle{empty}

\input{abstract}
\input{1-Introduction}
\input{2-Background}

\input{3-FrictionnalMPM}

\input{4-Experiments}

\input{5-Conclusion}


\section*{Acknowledgments}
This work has received support from the French National Research Agency, through the INEXACT project (ANR-22-CE33-0007-01) and under the France 2030 program with the references Organic Robotics Program (PEPR O2R) and “PR[AI]RIE-PSAI” (ANR-23-IACL-0008).
The European Union also supported this work through the ARTIFACT project (GA no.101165695) and the AGIMUS project (GA no.101070165).

\bibliographystyle{IEEEtran}
\bibliography{references}

\end{document}

%% file: macros.tex
\definecolor{pastelblue}{rgb}{0.68, 0.78, 0.81}

%% file: abstract.tex
\begin{abstract}
Frictional contact remains a core bottleneck for the Material Point Method (MPM), from reliable contact detection to enforcing the frictional contact laws (non-penetration, Coulomb friction, and maximum dissipation). We introduce a frictional-contact pipeline for implicit MPM that couples sub-cell detection to a globally-coupled, penalty-free resolution. Contacts are localized with particle-centric geometric proxies, then cast as a Nonlinear Complementarity Problem (NCP) over contact impulses and solved with an Alternating Direction Method of Multipliers (ADMM) scheme that reuses the implicit MPM linearization already factored for the internal mechanics. The method integrates seamlessly into the implicit MPM loop and is agnostic to material laws, interpolation functions, and transfer schemes. We validate it on an analytical stick-slip benchmark and demonstrate it across a broad range of MPM-based simulation scenarios relevant to robotics.

\end{abstract}

%% file: 1-Introduction.tex
\section{Introduction}
\label{sec:introduction}

Accurately handling contact in deformable simulation is essential for robotics systems, where robots make and break contact to move or manipulate objects~\cite{TeschnerColliSoft2005, CoevoetOptimContact2017, SantinaContactControlPWC2018, menager2024learning}. 
While the Finite Element Method (FEM) supports soft-body simulation~\cite{articleSorotoki, Faure2011SOFAAM}, large deformations, remeshing, and path-dependent plasticity remain challenging. The Material Point Method (MPM) offers a compelling alternative. 
MPM mixes an Eulerian and a Lagrangian view of matter: the Lagrangian view represents each body as a set of material points (particles), while the Eulerian view advances dynamics on a background computational grid. Particle states are transferred to the grid, where the equations of motion are solved. The updated grid state is then mapped back to particles. This dual view enables large deformation together with complex constitutive behavior without remeshing~\cite{SULSKY1994MPM, Jiang2016TheMP} and has been leveraged in recent robotics works for differentiable control and learning~\cite{Chainqueen2019, huang2021plasticinelab, ManiSkill22023}.

Yet, robust frictional contact remains a bottleneck for MPM. Detection is often performed on grid nodes: when two materials contribute mass to the same grid node, a contact is inferred, and a normal is built from mass/volume gradients~\cite{BardenhagenMPM2001, Huang2011MPM, HU2003, NEZAMABADI2015}. 
This approach is robust, but intrinsically diffuse at the cell scale. 
Boundary-aware alternatives improve geometry - e.g., particle-boundary level-set~\cite{LIU2020, SONG2025}, cell/sub-cell integration near interfaces~\cite{SONG2025}, boundary material points~\cite{kakouris2024, Han2019}, hybrid MPM / discrete elements scheme~\cite{Chen2022, Liu2018, Yue2018} -- but typically add modeling layers or penalties, and are frequently coupled to explicit stepping. 
More recently, barrier-based formulations tightly couple contact with the implicit solver through global optimization and barrier energies~\cite{LI2022114350}. 
While highly robust, these approaches rely on dedicated optimization pipelines with regularized barrier/contact potentials and barrier parameters, rather than an independent contact-resolution layer. On the resolution side, classical node-based approaches enforce no-penetration and no-slip implicitly by sharing the same interpolation space at a standard grid node~\cite{BardenhagenMPM2001}. 
Many methods remain local and sequential: a normal reaction from velocity projection or penalty is computed nodewise, then tangential forces are calculated using Coulomb law~\cite{SONG2025, kakouris2024, Homel2017, NEZAMABADI2015}. 
These methods require penalty tuning and ignore global coupling between contacts. Recent convex formulations~\cite{zong2024convex, yu2025convex} restore global structure but approximate the original complementarity model and introduce regularization parameters.

To overcome these limitations, we target both collision detection and collision resolution. 
For collision detection, we localize contacts at a sub-cell scale from particle-centric, geometry-aware proxies.
For collision resolution, we cast frictional contact as a Nonlinear Complementarity Problem (NCP) on contact impulses and solve it with an Alternating Direction Method of Multipliers (ADMM)-based scheme. 
We implement the pipeline inside an implicit MPM loop to enable large time steps and to reuse the linearization already factored for internal mechanics and constraints. 
Contact remains a separate resolution layer that reuses the implicit MPM admittance without modifying the underlying solver, and is independent of the collision detector, interpolation and transfer scheme, and constitutive model.
The NCP formulation and the ADMM solver follow the rigid-body and FEM treatments of~\cite{Carpentier2024FromCT, menager2025differentiable}. Our specific contribution is their coupling to implicit MPM, namely the sub-cell particle-centric detection, the barycentric deposition of contacts onto the grid, and the assembly of the Delassus operator from the reused grid admittance.
We evaluate eight scenes spanning MPM-rigid and MPM--MPM stick-slip, nonlinear materials, complex geometries, and kinematic manipulation, together with analytical validation, local baselines, timing, and stability.

%% file: 2-Background.tex
\section{Background}
\label{sec:background}

This section fixes the notation used throughout the paper. We recall the continuum equations of motion, the Material Point Method (MPM) discretization coupling Lagrangian particles to an Eulerian grid, and the implicit time-stepping scheme whose admittance matrix is later reused for contact. We then summarize our transfer, interpolation, and material choices, and introduce the contact quantities that connect the grid update to the frictional problem of \Cref{sec:frictionalMPM}. We refer the reader to~\cite{Jiang2016TheMP} for a comprehensive treatment of MPMs.

\subsection{Continuum model}
\label{subsec:continuum_model}

Let $\Omega_0$ be the reference configuration (undeformed body at $t=0$), and let $\bm X \in \Omega_0$ be a material point. 
The current position of the material point is $\bm x(\bm X, t)$, with deformation gradient $\bm F = \frac{\partial \bm x}{\partial \bm X}$ and volume variation $J = \det \bm F$. Balance of linear momentum in the reference frame is written as:
\begin{equation}
\label{eq:momentum}
\rho_0 \ddot{\bm x} = \nabla_{\bm X} \cdot \bm P + \rho_0 \bm f,
\end{equation}
where $\rho_0$ is the reference density, $\bm f$ the body force, and $\bm P$ the first Piola-Kirchhoff stress. For hyperelastic material with energy $W$, $\bm P = \frac{\partial W}{\partial \bm F}$. Boundary conditions complete the model. In MPM, we discretize the weak form of \eqref{eq:momentum}. 
This leads to the particle sums that appear in the formulas below.  

\subsection{MPM: mixing Lagrangian particles and an Eulerian grid}
\label{subsec:lagrange_euler}

MPM combines two representations of the same body. The Lagrangian view stores material history on a finite set of particles $p$, carrying mass $m_p$, volume $V_p$, position $\bm x_p$, velocity $\bm v_p$, deformation gradient $\bm F_p$, and possible internal variables (e.g., relative to plasticity). The Eulerian view uses a background grid as a computational backend to solve the momentum equation. At each simulation step, particles transfer mass and momentum to nearby grid nodes through interpolation. The grid then performs the time update, and the updated grid state is interpolated back to the particles. 

Let $N_i(\bm x)$ be the interpolation function of grid node $i$, $N_{ip} = N_i(\bm x_p)$ and $\nabla N_{ip} = \nabla N_i (\bm x) |_{\bm x_p}$. Let $n$ denote the current time-step index. The particle-to-grid (P2G) transfer accumulates nodal mass and momentum on the grid, 
\begin{align}
\label{eq:mass_momentum}
m_i = \sum_p m_p N_{ip},  \\
\bm p_i = \sum_p m_p \bm v_p N_{ip}, \nonumber
\end{align}
and computes internal nodal forces from particle stresses based on virtual work 
\begin{equation}
\label{eq:fint}
\bm f_i^{int} = - \sum_p V_p \bm P_p \nabla N_{ip}.
\end{equation}
External forces $\bm f^\text{ext}$ (e.g., gravity) are either added on the grid or on the particles and transferred to the grid using an interpolation function. 

After the grid update (see \Cref{subsec:implicit_scheme}), grid-to-particle (G2P) interpolates velocities and positions back, 
\begin{align}
\label{eq:G2P}
\bm v_p^{n+1} = \sum_i N_{ip} \bm v_i^{n+1}, \\
\bm x_p^{n+1} = \bm x_p^n + \Delta t \bm v_p^{n+1}, \nonumber
\end{align}
and advances the deformation gradient with the grid velocity gradient, 
\begin{equation}
\label{eq:def_grad}
\bm F_p^{n+1} = (\bm I + \Delta t \sum_i \bm v_i^{n+1} \otimes \nabla N_{ip}) \bm F_p^n.
\end{equation}

Stresses are then recomputed at particles from $\bm F_p^{n+1}$.

\subsection{Implicit time stepping and the admittance matrix}
\label{subsec:implicit_scheme}

We adopt an implicit update on the grid for stability and larger time steps. We take a single linearization of the discrete equations over a time step $\Delta t$ (a semi-implicit step), which yields a symmetric positive-definite system for the stacked nodal velocity increment $\Delta \bm v = \bm v^{n+1} - \bm v^n \in \mathbb{R}^{3n_g}$
\begin{equation}
\label{eq:implicit}
\underbrace{(\bm M + \Delta t \bm D + \Delta t^2 \bm K )}_{\bm A}\Delta \bm v = \underbrace{\Delta t (\bm f ^\text{ext} - \bm f^\text{int})}_{\bm b} 
\end{equation}
where $n_g$ is the number of grid nodes, $\bm M = \text{diag}(m_i)$ is the lumped mass, $\bm D$ is the damping matrix (computed here as Rayleigh damping), and $\bm K$ the tangent stiffness arising from the variation of internal forces. $\bm A^{-1}$ is the admittance matrix we will reuse for contact solving. Solving $\bm A \Delta \bm v = \bm b$ gives $\bm v^{n+1} = \bm v^n + \Delta \bm v$. Additional non-contact constraints may be incorporated through Lagrange multipliers.

The stiffness matrix $\bm K$ is computed from \eqref{eq:fint}. A perturbation of nodal velocities perturbs $\bm F_p$ via the grid velocity gradient, which in turn perturbs $\bm P_p$. Collecting contributions over particles yields $\bm K$ as a sparse block matrix whose entries depend on $\frac{\partial \bm P_p}{\partial \bm F_p}$, $\bm F_p$, and $\nabla N_{ip}$.

\subsection{Transfers, interpolation, and materials}
\label{subsec:transfer_interp_material}

Basic transfers can be affected by dissipation and cell-crossing artifacts. Several approaches exist to improve stability. Affine Particle-In-Cell (APIC)~\cite{jiang2015affine} augments particles with an affine velocity that preserves angular momentum and reduces numerical diffusion. Moving Least Squares MPM (MLS-MPM)~\cite{Hu2018MLSMPM} reconstructs grid and particle velocities using moving least squares to handle strong distortions.

For interpolation, higher-order B-spline interpolation~\cite{Gan2017BSpline} enlarges the support and yields smoother gradients. More recently, CK-MPM~\cite{liu2025ck} introduced a $\mathcal{C}^2$ compact kernel and a dual-grid framework that mitigates cell-crossing with reduced numerical diffusion, while remaining compatible with APIC/MLS, offering a stable and efficient alternative to wide B-spline supports. 

Regarding constitutive behavior, we support standard elastic and hyperelastic laws (linear, St. Venant-Kirchhoff, co-rotational, and Neo-Hookean) as well as elasto-plastic extensions.  Plasticity is handled at particles by the multiplicative split $\bm F = \bm F_e \bm F_{pla}$: after the elastic trial update, a plastic projection (e.g., SVD-based projection) enforces the yield constraints and updates $\bm F_{pla}$. Stresses $\bm P(\bm F_e)$ then follow from the post-projection state. 
This particle-level mechanism allows considering plasticity while keeping the grid unchanged. 

\begin{figure*}[!ht]
\centering
\resizebox{0.63\textwidth}{!}{
\includegraphics{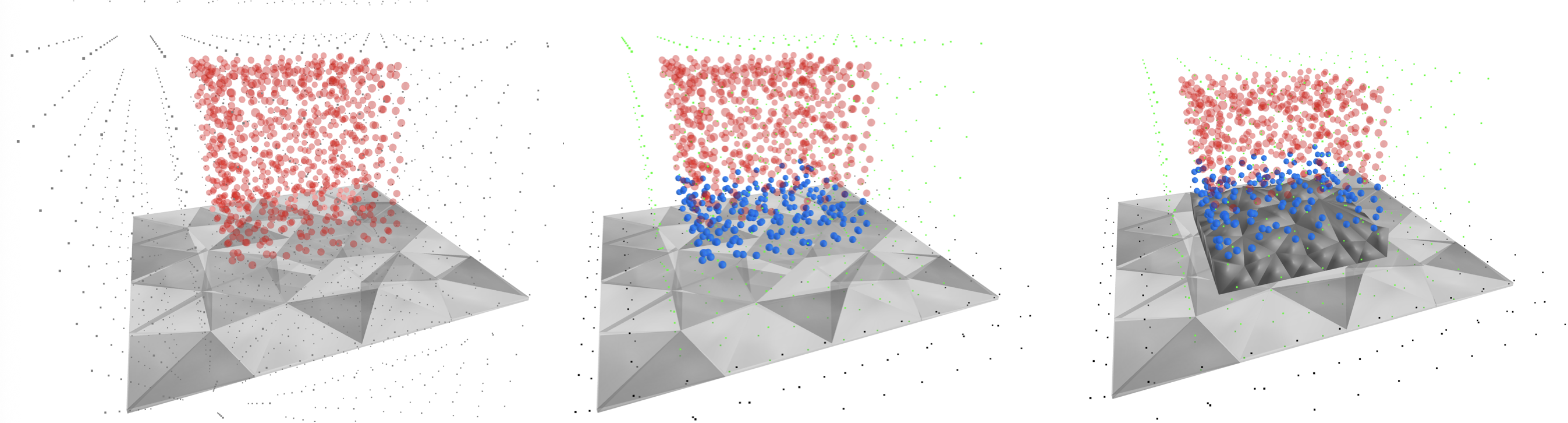}
}
\caption{\textbf{Particle-centric contact detection and grid-based resolution.} \textbf{Left:} A block is discretized into particles (red) and falls onto a planar surface. Gray dots represent MPM grid nodes. \textbf{Center:} A shared-grid broad phase identifies candidate particle-tetrahedron pairs for the narrow phase (blue). \textbf{Right:} Deformed tetrahedral proxies (dark gray) are used to localize contact points and frames. Contacts are barycentrically transferred to the grid to assemble the contact Jacobian, then resolved as a global NCP using ADMM. The resulting impulses correct grid velocities before G2P.}
\label{fig:frictional_detection_resolution}
\vspace{-0.5cm}
\end{figure*}

\subsection{Contact notation}
\label{subsec:contact_notation}

To couple contact consistently with the implicit step, we map grid velocities to relative velocities at contact points via a contact Jacobian $\bm H_c$. For a set of contact frames $c$ with normal $\bm n_c$ and tangents $\bm t_{c,1}$, $\bm t_{c,2}$, the stacked contact velocity is $\bm \sigma_c = \bm H_c \bm v$, and the contact impulse is $\bm \lambda_c = [\lambda_{c,N}, \bm \lambda_{c,T}]$ constrained by a Coulomb cone. 
The Delassus operator \mbox{$\bm G = \bm H_c \bm A^{-1} \bm H_c^\top$} maps contact impulses to contact velocities. Advancing the grid without contact yields a free contact velocity $\bm g = \bm H_c \bm v^\text{free}$, contact then corrects it by solving a friction problem in $\bm \lambda_c$. 
The following section formalizes this as a nonlinear complementarity problem.

%% file: 3-FrictionnalMPM.tex
\section{Frictional contact for implicit MPM}
\label{sec:frictionalMPM}

In this section, we present our frictional contact model, which separates \emph{where} contacts are detected from \emph{how} they are resolved. Detection is particle-centric and geometry-aware, giving sub-cell localization consistent with the deformed shape (\Cref{subsec:contact_particles}); resolution is grid-based, coupling each contact to nearby nodes (\Cref{subsec:grid_deposition}) and enforcing Coulomb friction as a single global nonlinear complementarity problem (NCP) solved with a specialized ADMM~\cite{Carpentier2024FromCT}. We close with compatibility and multi-object scenes. \Cref{fig:frictional_detection_resolution} illustrates this pipeline.

\subsection{Contact discretization on particles}
\label{subsec:contact_particles}

\smallskip
\noindent\textbf{Broadphase.} We select potentially interacting pairs using a uniform background grid shared across objects. Overlapping cells yield candidate particle pairs (or particle-tetrahedron pairs for mesh obstacles). This reduces complexity and is independent of the MPM grid used for dynamics.

\smallskip
\noindent\textbf{Narrowphase.} Contacts are localized using deformed particle-centered geometric proxies. Each particle carries a tetrahedron whose vertices are advected by the particle’s deformation gradient. We query tetrahedron-tetrahedron proximity using a variation of the GJK algorithm~\cite{montaut2024gjk}. For each detected contact, we compute:
\begin{itemize}
\item a contact point $\bm x_c$ on the deformed primitives;
\item a unit normal $\bm n_c$;
\item an orthonormal contact frame $\bm R_c = [\bm t_{c,1}, \bm t_{c,2}, \bm n_c]$.
\end{itemize}

This particle-centric, geometry-aware detection provides sub-cell localization, in contrast to node-wise grid detection, and ensures the contact geometry is consistent with the actual deformable shape. 

\subsection{Grid deposition of contact frames}
\label{subsec:grid_deposition}

Contacts are enforced on the grid, so each contact must couple to nearby grid nodes. 
We process in two steps.

\smallskip
\noindent\textbf{Step 1 -- Barycentric lift to tetrahedron vertices.} A contact point $\bm x_c$ lying inside a deformed tetrahedron with vertices $\{\bm x_k\}_{k=1…4}$ is expressed with barycentric weights $\{\phi_k\}$ such that $\bm x_c = \sum_k \phi_k \bm x_k$. To ensure active nodal support, we represent the contacting particle by four vertex proxies at $\{x_k\}$. Proxy $k$ carries barycentric weight $\phi_k$. P2G deposits contact through these proxies, rather than the particle centroid, thereby avoiding empty-cell contacts and ensuring deposition only on nodes with mass.

\noindent\textbf{Step 2 -- Vertex-to-grid interpolation.} 
Each vertex contributes to grid nodes $i$ through the same interpolation $N_i(\cdot)$ as in the P2G/G2P pipeline. Stacking nodal velocities into $\bm v \in \mathbb{R}^{3n_g}$, the contact Jacobian row block $\bm H_{c, j} \in \mathbb{R}^{3 \times 3n_g}$ maps nodal velocities to contact-frame $j$ relative velocity:
\begin{equation}
\label{eq:contact_vel}
\bm \sigma_{c,j} = \begin{bmatrix} v_{t,1} \\ v_{t,2} \\ v_n \end{bmatrix} = \bm H_{c,j} \bm v,
\end{equation}
with
\begin{equation}
\label{eq:Hc}
\bm H_{c,j} = \sum_{k=1}^4 \phi_k \sum_{i \in \mathcal{N}(k)} \begin{bmatrix} \bm t_{c,1} \\ \bm t_{c,2} \\ \bm n_c \end{bmatrix}_j \otimes N_i(\bm x_k) \bm e_i^\top,
\end{equation}
where $\mathcal{N}(k)$ are grid nodes influencing vertex $k$, and $\bm e_i$ selects node $i$’s 3-DOF block. The global Jacobian matrix is obtained by stacking all contact Jacobian rows. This barycentric deposition guarantees active nodal support and preserves geometric locality. 

\begin{figure*}[!h]
\centering
\resizebox{0.63\textwidth}{!}{
\includegraphics{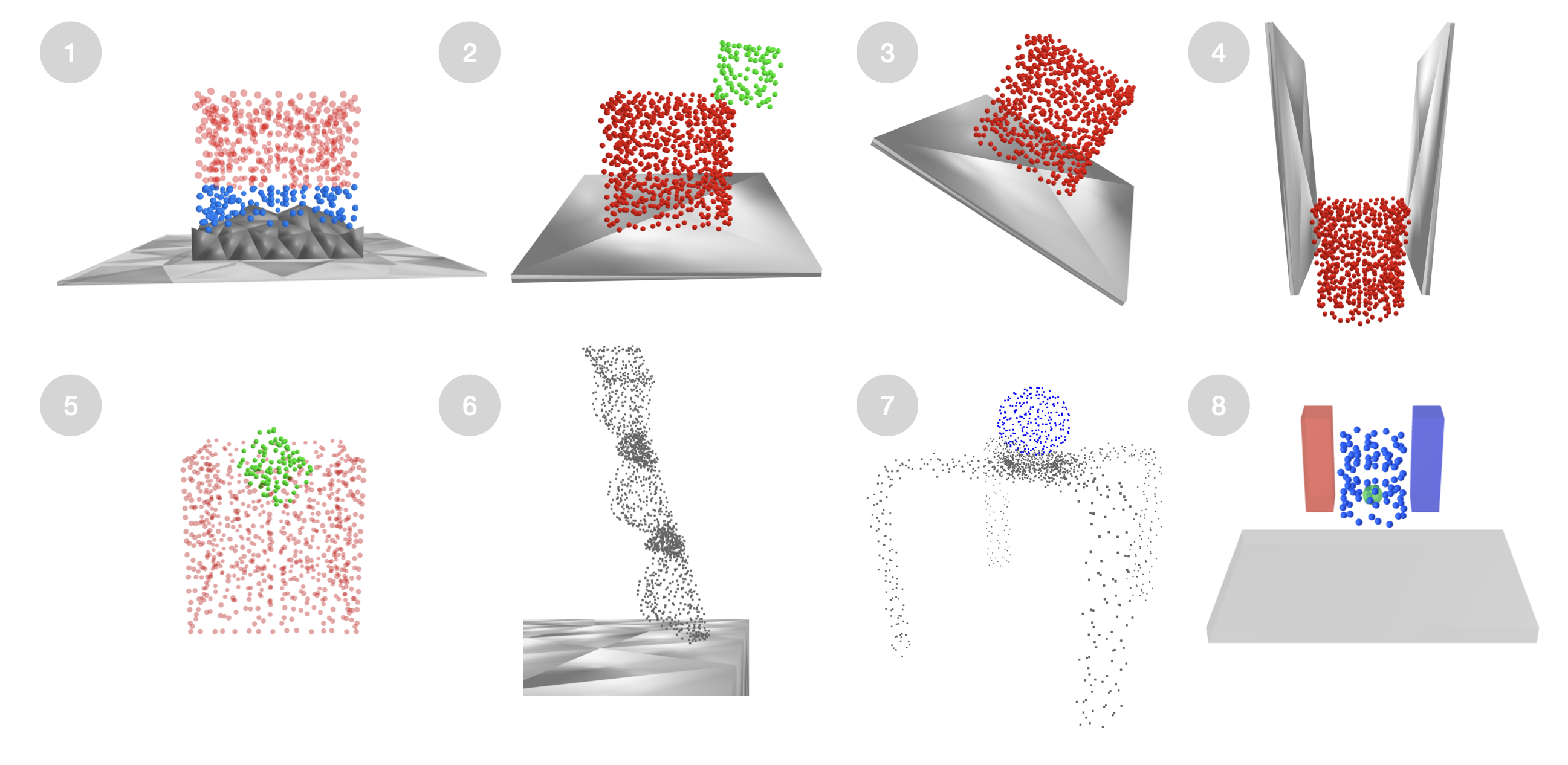}
}
\caption{\textbf{Illustration of the eight simulation scenes considered in this paper.} (1) a deformable cube falling onto a fixed plane; (2) interaction between two MPM bodies; (3) a deformable cube sliding on an inclined plane; (4) a plastic cube sliding between two fixed walls; (5) a cube entering a fluid-like medium; (6) a soft robotic finger rubbing against a fixed plane; (7) a sphere impacting a quadrupod soft robot; and (8) a kinematic parallel-jaw gripper lifting a deformable object.}
\label{fig:experiments}
\vspace{-0.5cm}
\end{figure*}

\subsection{Global frictional contact as an NCP on the grid}
\label{subsec:NCP_on_grid}

\smallskip
\noindent\textbf{Frictional contact as an NCP.} Contact modeling involves various complementarity physical principles~\cite{Lidec2024Contact}. First, the so-called Signorini condition provides a complementarity constraint $0 \leq \lambda_{c, N} \perp \sigma_{c, N} \geq 0$, where $_N$ stands for the normal component of the contact force and contact point velocity. The Signorini condition ensures that the normal force is repulsive, the bodies do not interpenetrate, and no power is injected by the normal forces. Secondly, the maximum dissipation principle combined with the frictional Coulomb law $\| \bm \lambda_{c, T}\|_2 \leq \mu \lambda_{c,N}$ of friction $\mu$, states that the tangential component of the contact forces $\bm \lambda_{c, T}$ maximizes the power dissipated by the contact. These principles are equivalent to the following NCP:
\begin{align}
\label{eq:NCP_problem}
     \mathcal{K}_\mu &\ni \bm \lambda_c \perp \bm \sigma_{c} + \Gamma_\mu(\bm \sigma_c) \in \mathcal{K}_\mu^*, \\
    \bm \sigma_{c} &= \bm G \bm \lambda_c + \bm g, \nonumber
\end{align}
where $\bm G$ is the Delassus operator mapping contact impulses to contact-point velocities, $\bm g$ is the free contact velocity obtained when $\bm \lambda_c = \bm 0$, $\mathcal{K}_{\mu}$ is a second-order cone with aperture angle $\arctan(\mu)$, $\mathcal{K}_{\mu}^*$ is the dual cone of $\mathcal{K}_{\mu}$, and $\Gamma_\mu(\bm \sigma_c) = [0, 0, \mu \| \bm \sigma_{c,T}\|_2]$ is the De Saxcé correction enforcing the Signorini condition~\cite{de1998bipotential,Lidec2024Contact}. This NCP can be solved for $\bm \lambda_c$ using different optimization methods, such as Projected Gauss-Seidel methods (PGS)~\cite{acary2018solving, Lidec2024Contact} or ADMM~\cite{acary2018solving, Carpentier2024FromCT}.

\smallskip
\noindent\textbf{Resolution on the grid.} In the case of the MPM, the implicit step produces the admittance matrix $\bm A^{-1}$ and the free grid velocity \mbox{$\bm v^\text{free} = \bm v^n + \bm A^{-1} \bm b$}, so that the free contact velocity is $\bm g = \bm H_c \bm v^\text{free}$ and the Delassus operator takes the form \mbox{$\bm G = \bm H_c \bm A^{-1} \bm H_c^\top$}. The solution $\bm\lambda_c$ of the NCP then corrects the grid velocities by
\begin{align}
\label{eq:contact_correction}
\bm v^{n+1} = \bm v^\text{free} + \bm A^{-1} \bm H_c^\top \bm \lambda_c, \\
\bm \sigma_c = \bm H_c \bm v^{n+1} = \bm g + \bm G \bm \lambda_c \nonumber.
\end{align}
Through $\bm G$, all contacts are coupled by the shared admittance matrix and resolved \emph{simultaneously} rather than one contact at a time. Crucially, the factorization of $\bm A$ computed for the implicit step~\eqref{eq:implicit} is reused to resolve contacts. We solve~\eqref{eq:NCP_problem} with an ADMM scheme (as used in rigid-body and FEM simulation~\cite{Carpentier2024FromCT, menager2025differentiable}), alternating a linear solve and a proximal projection onto the friction cone, with stopping criteria based on the primal/dual residuals of the NCP. We refer to~\cite{Carpentier2024FromCT} for more details.

\subsection{Compatibility and multi-object simulation}
\label{subsec:compatibility}

The global approach is summarized in \Cref{alg:mpm_frictionnal}. It is agnostic to constitutive models and transfer/interpolation. It only requires consistent P2G/G2P and the admittance matrix. 

For multi-object simulation, we maintain one MPM grid per object, with a shared origin and spacing. The contact Jacobian concatenates per-object blocks, and the Delassus contains cross-terms that couple objects through shared contacts. For a prescribed rigid boundary, the free relative velocity at
contact $j$ becomes $\bm g_j=\bm H_{c,j}\bm v^{\mathrm{free}}
-\bm R_{c,j}^{\top}\bm v_b(\bm x_{c,j})$, where $\bm v_b$ is the prescribed boundary velocity and $\bm R_{c,j}$ is the contact frame. The boundary contributes no admittance block and receives no dynamic correction. $\bm v_{b}(\bm x_{c,j})=\bm 0$ for fixed boundaries.

\begin{algorithm}[h]
\caption{MPM simulation with frictional contacts.}
\label{alg:mpm_frictionnal}
\SetAlgoLined
\textbf{Initialize:}  a set of particles $\{p\}$ per object, a background grid per object\;
\For{$k = 0, 1, \ldots, K$}{
    \textbf{Step 1}: P2G and assembly of $\bm A$, $\bm b$; factorize $\bm A$ (once per object)\; 
    \textbf{Step 2}: Free step: $\bm v^\text{free} = \bm v^n + \bm A^{-1} \bm b$\; 
    \textbf{Step 3}: Detection on particles and creation of the contact frames $[\bm x _c , \bm R_c]$ \; 
    \textbf{Step 4}: Build $\bm H_c$ by barycentric lift to tetrahedron vertices, then vertex-to-grid deposition \; 
  \textbf{Step 5}: ADMM solve of  \eqref{eq:NCP_problem} using  $G= \bm H_c \bm A^{-1} \bm H_c^\top$ and $\bm g = \bm H_c \bm v^\text{free}$\;
\textbf{Step 6}: Correct grid velocity $\bm v^{n+1} = \bm v^\text{free} + \bm A^{-1} \bm H_c^\top \bm \lambda_c$ \; 
\textbf{Step 7}: G2P update of particles\;}
\end{algorithm}

%% file: 4-Experiments.tex
\section{Experiments}
\label{sec:experiments}

We evaluate our method along four axes: (i) physical correctness against analytical references, (ii) fidelity of frictional stick-slip behavior compared to local node-based baselines, (iii) robustness across materials, interpolation, and transfer schemes, and (iv) applicability to robotics-relevant scenarios involving complex geometries, nonlinear materials, and kinematic manipulation. To this end, we consider eight representative scenes, illustrated in~\Cref{fig:experiments}, ranging from MPM-rigid and MPM-MPM contact to soft-robot interaction. The accompanying video illustrates all scenarios.

Unless stated otherwise, we use quadratic B-spline interpolation, basic P2G/G2P transfer, and a Neo-Hookean material. We additionally evaluate linear, St.~Venant-Kirchhoff, and co-rotational elasticity, as well as APIC, CK-APIC, and CK-MLS transfer schemes.

\smallskip
\noindent\textbf{Setup and implementation.} Particles are generated from tetrahedral object meshes, with one particle initialized at each tetrahedron barycenter and the tetrahedron defining its particle-centered geometric proxy. Unless stated otherwise, $\Delta t \in [10^{-3}, 10^{-2}]\,\mathrm{s}$, and the grid spacing is selected such that each active cell initially contains approximately six to ten particles. 

Collision detection uses Coal~\cite{coalweb} on the deformed tetrahedral proxies during the narrow phase. Frictional contact is solved using the ADMM formulation of~\cite{Carpentier2024FromCT} implemented within the Pinocchio library~\cite{carpentier2015pinocchio,carpentier2019pinocchio}, with a sparse linear backend that reuses the implicit MPM factorization. We use an absolute tolerance of $10^{-6}$ and a maximum of $1000$ ADMM iterations per simulation step. 
The implementation is written in Python and will be released upon paper acceptance.

\smallskip
\noindent\textbf{Experiment 1 -- Single- and multi-object contact.} 
We first evaluate contact involving both rigid boundaries and multiple deformable objects. 
In the first scene, a deformable block falls onto a fixed floor (\Cref{fig:experiments}.1). Across nine material, interpolation, and transfer configurations, the mean number of active contacts ranges from $46.1$ to $95.4$, and the mean ADMM iteration count ranges from $274$ to $613$. Depending on the configuration, $71.85\%$ to $85.11\%$ of the steps reach a final residual below $10^{-6}$, with a maximum residual of approximately $10^{-4}$. The residual measures the NCP primal-dual optimality. Steps reaching the iteration cap correspond to the densest contact sets but still satisfy non-penetration constraint to visual tolerance. The implicit run on this scene completes without measured penetration (\Cref{fig:experiments}.1, cf.\ Experiment~4).
The second scene considers MPM-MPM interaction (\Cref{fig:experiments}.2). A deformable block falls onto a second block resting on a fixed floor, with several initial lateral offsets. The Delassus operator $\bm G$ contains cross-object terms and therefore couples the normal and tangential responses of both deformable bodies without penalty tuning. 
Together, these scenes validate the complete contact pipeline.

\begin{figure*}[!h]
\centering
\resizebox{0.5\textwidth}{!}{
\includegraphics{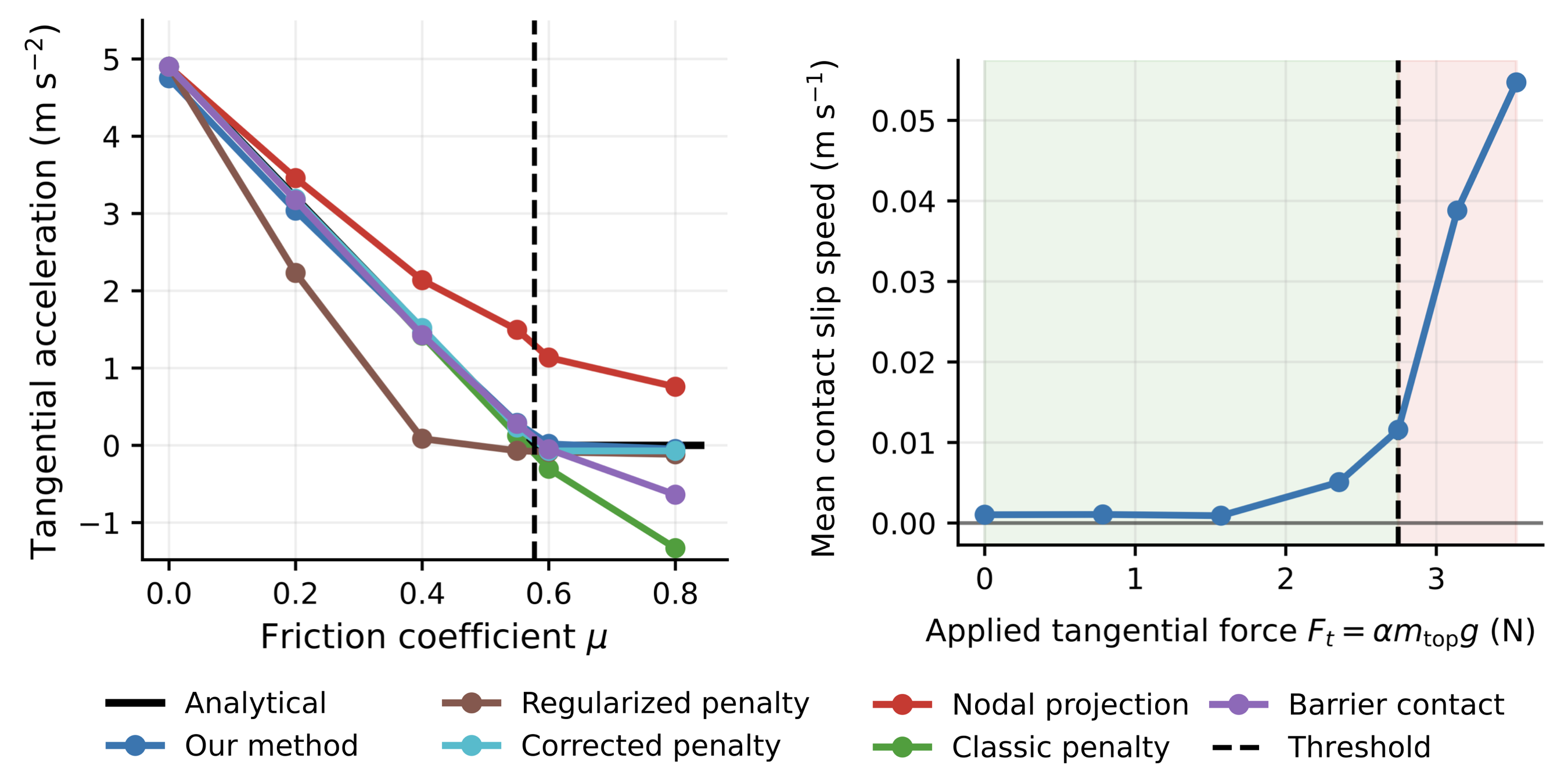}
}
\caption{\textbf{Stick-slip benchmarks.}
Left (MPM-rigid): fitted tangential COM acceleration as a function of friction for the different contact methods; the analytical response is shown in black. The corrected-penalty baseline is the gravity-corrected variant, which injects the analytically known normal load and is therefore problem-specific. Right (MPM-MPM): mean tangential contact slip speed as a function of the applied tangential load.}
\label{fig:stick-slip-benchmarks}
\end{figure*}

\begin{table*}[t]
  \caption{\textbf{Accuracy, penetration, and wall-clock runtime for the inclined-plane benchmark.} Lower values are better. Maximum penetration is measured as the largest negative signed distance of the contact point to the analytical plane over the complete trajectory. Runtime is averaged over the six friction-coefficient runs.}
  \label{tab:inclined_plane}
  \centering
  \begin{adjustbox}{max width=\textwidth}
    \begin{tabular}{r|cccc}
      \toprule
      \textbf{Method} & \textbf{RMSE} $(\mathrm{m\,s^{-2}})$ & \textbf{Relative $\ell_2$ error} & \textbf{Max. penetration} $(\mathrm{m})$ & \textbf{Mean runtime per run} $(\mathrm{s})$ \\
      \midrule
      \rowcolor{pastelblue}
      Our method & 0.1010 & 0.0409 & 0 & 4.8870 \\
      Nodal projection & 0.8083 & 0.3270 & 0 & 4.2009 \\
      \rowcolor{pastelblue}
      Classical penalty & 0.5600 & 0.2266 & 0.00620 & 4.3200 \\
      Regularized penalty & 0.7165 & 0.2899 & 0.37110 & 4.3383\\
      \rowcolor{pastelblue}
      Corrected penalty & 0.0424 & 0.0171 & 0 & 4.2979 \\
      Local barrier force & 0.2647 & 0.1071 & 0 & 4.3146 \\
      \bottomrule
    \end{tabular}
  \end{adjustbox}
\end{table*}

\smallskip
\noindent\textbf{Experiment 2 -- Stick-slip transitions.} 
We study two controlled stick-slip scenarios. The first provides an analytical reference for MPM-rigid contact, while the second verifies that the same contact formulation captures the expected stick-slip transition in MPM-MPM interaction.

\smallskip
\noindent\textit{Inclined-plane benchmark.} 
A deformable cube is placed on a plane rotated by $-30^\circ$ around the $x$-axis (\Cref{fig:experiments}.3). We set $\theta = 30^\circ$, define the positive tangential direction downhill, and measure cube's center of mass (COM) tangential acceleration while contact is maintained. 

Let $g_0 = \| \bm g_{\mathrm{grav}}\|$ denote the magnitude of gravitational acceleration. Neglecting elastic transients and assuming sustained contact, the critical friction coefficient is \mbox{$\mu^\star = \tan \theta \approx0.577$}, and the analytical acceleration is 
\begin{equation}
\label{eq:analytical-stick-slip}
a_{\mathrm{ana}}(\mu) = 
    \begin{cases}
        g_0\left(\sin\theta-\mu\cos\theta\right), & \mu < \mu^\star,\\
        0, & \mu \geq \mu^\star.
    \end{cases}
\end{equation}

We compare our method against three families of \emph{local} contact-resolution strategies. These are chosen as a controlled ablation of the resolution layer: each shares our detection, transfer, and time integration, and differs only in how the frictional response is enforced, thereby isolating the effect of the global NCP resolution. A comparison against globally-coupled state-of-the-art formulations -- notably convex MPM contact~\cite{zong2024convex, yu2025convex} and barrier MPM--FEM coupling~\cite{LI2022114350} -- under matched tolerances is a natural next step, left to future work as it requires integrating separate solver stacks. The three local strategies are:
\begin{itemize}
    \item \textit{Nodal contact projection}: inspired by classical multi-mesh MPM contact schemes~\cite{PanMPMMultiMesh}, and specialized to a rigid analytical plane with basic transfer. Grid nodes influenced by particles near the plane are identified, and their velocities are locally projected onto the plane's Coulomb friction cone. 

    \item \textit{Penalty-based contact}: inspired by particle-boundary penalty formulations~\cite{kakouris2024}, and specialized to the analytical plane. We consider a classical penetration-based spring-damper penalty, a band-regularized variant activated before penetration, and a gravity-corrected variant that additionally uses the expected normal load. All use viscous tangential damping capped by the Coulomb bound and therefore regularize, rather than exactly enforce, static friction. The corrected variant injects the known gravity-induced normal load and is deliberately problem-specific.

    \item \textit{Barrier-based contact}: inspired by IPC-style barrier models~\cite{Li2020IPC}. We apply a local normal barrier force against the analytical plane together with Coulomb-capped tangential damping. This is not a complete IPC formulation, since it omits the global optimization procedure and barrier Hessian. 
\end{itemize}

All methods use the same material, grid spacing, initial state, and simulation horizon. For each parameterized method, a single parameter set is selected by coarse grid search over the six-point friction sweep and kept fixed for all $\mu$. Acceleration is obtained from the least-squares position fit \mbox{$x_t(t) = c + v_0 t + \frac 12 a_{\mathrm{fit}}t^2$}. We report \mbox{$\mathrm{RMSE} =\|\bm a_{\mathrm{fit}}-\bm a_{\mathrm{ana}}\|_2/\sqrt{6}$} and \mbox{$E_{\mathrm{rel}} =\|\bm a_{\mathrm{fit}}-\bm a_{\mathrm{ana}}\|_2/ \|\bm a_{\mathrm{ana}}\|_2$}. The results are shown in~\Cref{fig:stick-slip-benchmarks} left and~\Cref{tab:inclined_plane}.

The gravity-corrected penalty obtains the lowest global error, but explicitly uses the analytically gravity-induced normal load and is therefore not applicable to general contact configurations. Among the methods without this problem-specific information, ours achieves the lowest RMSE and relative error, with no measured penetration. Its measured runtime is approximately $16\%$ above the fastest method in this benchmark.
Nodal projection underestimates static friction and predicts substantial downhill acceleration above the analytical threshold. The classical penalty exhibits negative high-friction acceleration and finite penetration, illustrating its sensitivity to stiffness and damping. The regularized penalty is strongly biased near the transition and produces the largest penetration. The local barrier follows the sliding regime more closely but overshoots into negative acceleration at high friction. 

\noindent\textit{MPM-MPM benchmark.} We next consider a small deformable cube resting on a larger deformable cube (\Cref{fig:experiments}.2). After an initial settling phase, a constant tangential force $F_t(\alpha) = \alpha m_{\mathrm{top}} g_0$ is applied to the upper cube, with $m_{\mathrm{top}}$ the mass of the upper cube and $\alpha$ a dimensionless loading factor. For $\mu = 0.35$, the ideal quasi-static stick-slip threshold is \mbox{$F_t^\star = \mu m_{\mathrm{top}} g_0$}, or equivalently $\alpha^\star = \mu$.

We report the mean tangential slip speed $\langle\|\bm\sigma_{c,T}\|_2\rangle$ over  active MPM-MPM contacts after the initial transient. The results are reported in~\Cref{fig:stick-slip-benchmarks} right. Below the analytical threshold, the slip speed remains small, indicating macroscopic sticking. Above it, the slip speed increases sharply and the upper cube undergoes visible translation relative to the lower body. Small nonzero relative velocities below the threshold arise from elastic shear and discrete contact-set changes. Persistent contact tracking and temporally smoother contact-frame updates could further sharpen the transition.

\smallskip
\noindent\textbf{Experiment 3 -- Computational performance.} All timings are measured with rendering disabled on an Apple M3 Pro with 36~GB of unified memory under macOS. 
In the MPM-MPM stick-slip benchmark (\Cref{fig:experiments}.2), the contact set is sparse, with $14.5$ active contacts per step on average. The complete contact pipeline accounts for $3.8\%$ of the step time, whereas implicit MPM assembly and free-motion integration account for approximately $91\%$. Contact is therefore not the bottleneck in this regime. \Cref{tab:wall_clock_mpm_mpm} reports the contact timing breakdown.

\begin{table}[!h]
  \caption{Mean contact-pipeline timing for the sparse MPM-MPM benchmark. Rendering is disabled.}
  \label{tab:wall_clock_mpm_mpm}
  \centering
  \begin{adjustbox}{max width=\columnwidth}
    \begin{tabular}{r|cc}
      \toprule
      \textbf{Component} & \textbf{Time} $(\mathrm{ms})$ & \textbf{Full-step fraction} \\
      \midrule
      \rowcolor{pastelblue}
      Complete contact pipeline & 46.7 & 3.8\% \\
      Contact detection & 15.3 & 1.2\% \\
      \rowcolor{pastelblue}
      Contact Jacobian assembly & 16.6 & 1.3\% \\
      Delassus assembly & 10.6 & 0.9\%\\
      \rowcolor{pastelblue}
      ADMM solve & 2.8 & 0.2\% \\
      Contact correction & 1.3 & 0.1\% \\
      \bottomrule
    \end{tabular}
  \end{adjustbox}
\end{table}

The inclined-plane benchmark (\Cref{fig:experiments}.3) represents a much denser regime, reaching up to $667$ active contacts per step along the simulated trajectory. At this density, matrix assembly and the ADMM solve account for approximately $35\%$ of the total step time. By contrast, the local projection, penalty, and barrier baselines spend less than $1\%$ of the step time on contact, but do not solve a globally coupled frictional contact problem. 

To evaluate the potential of hardware acceleration, we implemented the algebraic pre-ADMM stage of this example in JAX and benchmarked it on a representative state with 667 active contacts. The median runtime was $1177.65$~ms for the Python/NumPy reference implementation, $154.03$~ms for JAX on CPU ($7.65\times$ speedup), and $12.55$~ms for double-precision JAX on an NVIDIA A100 ($93.86\times$ speedup), with host-to-device transfers included. These figures cover only the algebraic pre-ADMM stage and exclude COAL collision detection and the ADMM solve, which are not yet JAX-compatible and remain on the CPU. With the inputs already transferred to the GPU, the free-velocity solve, contact Jacobian assembly, and Delassus assembly achieve speedups of $10\times$, $1454\times$, and $90\times$, respectively.

These results show that the contact cost can be substantially reduced through JAX and GPU execution. Contact sparsification, warm-starting, matrix-free Delassus products, and GPU-compatible collision detection and ADMM are direct avenues for further acceleration.

\smallskip
\noindent\textbf{Experiment 4 -- Stability and proxy conditioning.} 
We evaluate the influence of the integration scheme in the falling-cube scene over a common simulation horizon of $T=0.2\,\mathrm{s}$. The implicit scheme remains stable and penetration-free for $\Delta t\in\{10^{-4},10^{-3},10^{-2}\}\,\mathrm{s}$. The explicit scheme remains stable up to $10^{-3}\,\mathrm{s}$ but fails through element inversion at $10^{-2}\,\mathrm{s}$, stopping after $12$ of $20$ steps with $\min\det(\bm F)=-0.245$. The implicit run completes all $20$ steps with $\min\det(\bm F)=0.979$ and no penetration. 

We additionally test the stability of $27$ positively oriented proxy configurations covering anisotropic stretching, shear, and flattening. 
Each proxy is placed near the plane $z=0$. Against exact analytical-plane queries, all expected contact states, depths, and normals are recovered, with a maximum depth error of $3.47\times10^{-18}\,\mathrm{m}$. Barycentric reconstruction remains accurate up to $\kappa_2(\bm F)\approx10^6$, with maximum coordinate and maximum relative momentum-conservation residuals of $5.03\times10^{-11}\,\mathrm{m}$ and $6.11\times10^{-16}$, respectively. In the two-wall simulation (\Cref{fig:experiments}.4), no inversion or deposition failure occurs. The minimum $\det(\bm F)$ is $0.5266$, the maximum condition number is $1.1804$, and the maximum edge-to-altitude ratio changes only from $5.686$ to $5.693$.

\smallskip
\noindent\textbf{Experiment 5 -- Complex materials and robotic manipulation.}
We finally consider nonlinear materials, complex geometries, and robotic manipulation. A plastic cube compressed between walls combines frictional dissipation and plastic flow (\Cref{fig:experiments}.4), while a cube entering a fluid-like medium interacts with a material carrying volumetric pressure but no shear memory (\Cref{fig:experiments}.5). The framework also handles sticking and sliding of an actuated soft finger (\Cref{fig:experiments}.6) and distributed MPM-MPM contact during the impact of a sphere on a quadrupod robot (\Cref{fig:experiments}.7).

\begin{figure}[!h]
\centering
\resizebox{0.5\linewidth}{!}{
\includegraphics{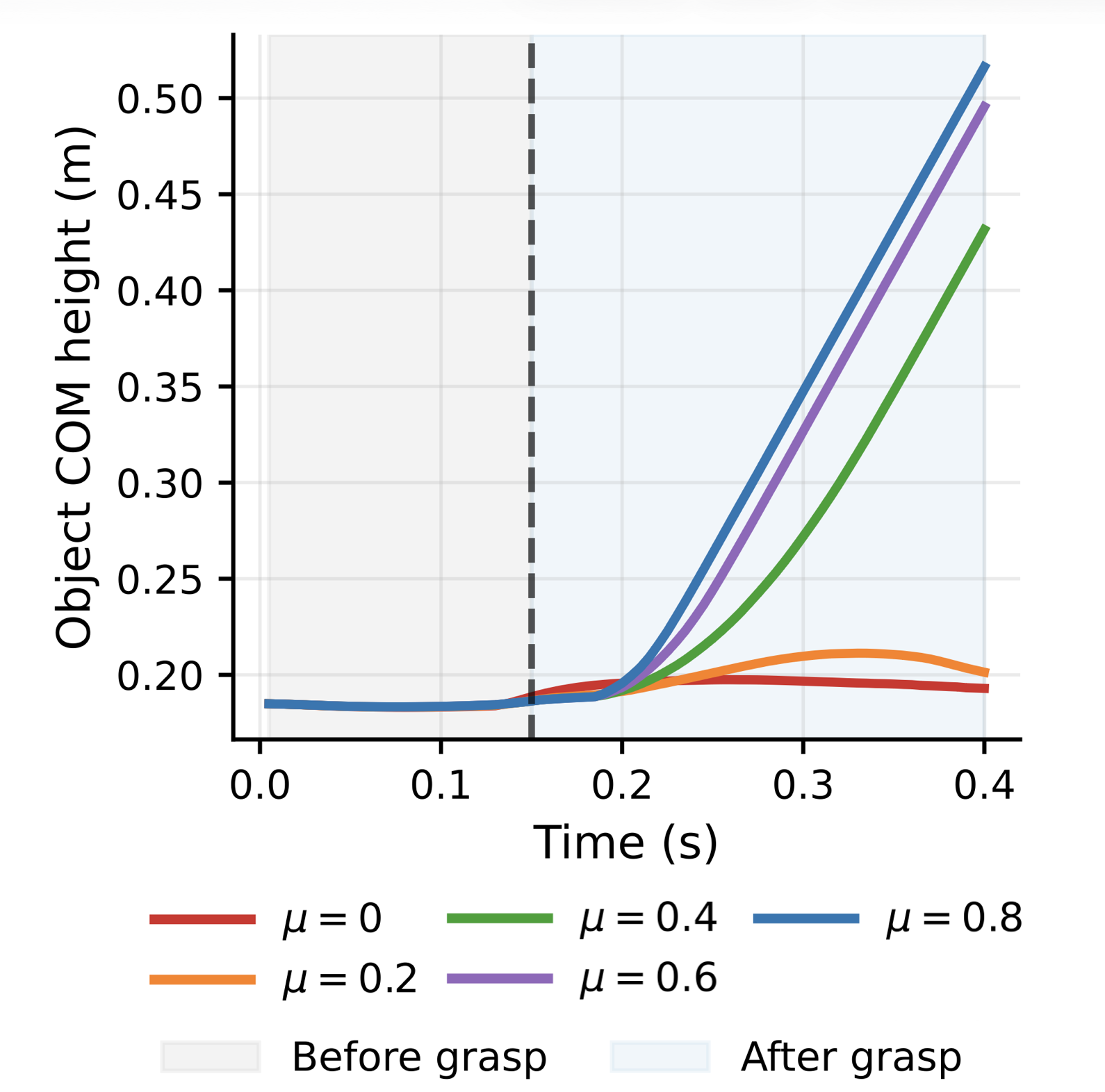}
}
\caption{Object COM height under the same prescribed kinematic grasp-and-lift trajectory for different friction coefficients. The vertical dashed line marks the beginning of the grasping phase. Larger friction yields sustained lifting.}
\label{fig:gripper_lift}
\end{figure}

Finally, a kinematic parallel-jaw gripper performs a prescribed grasp-and-lift motion on a deformable object (\Cref{fig:experiments}.8). The same trajectory is used for all friction coefficients, and the prescribed jaw velocities are included in the relative contact velocities. The gripper does not receive dynamic contact corrections. As shown in~\Cref{fig:gripper_lift}, low friction causes immediate slip, intermediate friction produces a partial lift, and sufficiently large friction yields sustained lifting. This example highlights the importance of accurately modeling friction in manipulation tasks. 

%% file: 5-Conclusion.tex
\section{Conclusion}
\label{sec:conclusion}

In this paper, we introduced a frictional-contact pipeline that integrates naturally with implicit MPM by separating \emph{where} contacts are detected from \emph{how} they are resolved. Contacts are detected at the sub-cell scale from particle-centric geometry, deposited onto the grid through the MPM interpolation, and resolved globally as a single nonlinear complementarity problem over contact impulses, reusing the admittance factorization already computed for the implicit step. Because contact is handled as a self-contained resolution layer, the pipeline remains agnostic to the collision detector, transfer and interpolation schemes, and constitutive model.
Across elastic and elasto-plastic scenes spanning sliding, stick-slip transitions, multi-contact, and complex geometries, the method delivers sub-cell contact localization and penalty-free Coulomb friction. On the analytical inclined-plane benchmark, it recovers the theoretical stick-slip response and, among local resolution strategies that do not inject a problem-specific normal load, attains the lowest error -- roughly $8\times$ below nodal projection -- without penalty tuning or observable penetration.
Future work therefore targets scalability -- through contact sparsification, temporally coherent detection, warm-starting, and GPU acceleration -- as well as broader validation and capabilities, including quantitative comparison against convex and barrier baselines, dynamic two-way rigid coupling, tighter integration with optimization-based contact solvers, and differentiable simulation for control and learning, in the vein of~\cite{de2018end,qiao2021differentiable,le2024leveraging,menager2025differentiable}.